\documentclass[10pt, a4paper]{article}

\usepackage{lrec-coling2024} 

\usepackage{lipsum}  

\hypersetup{
   breaklinks=true,   
   colorlinks=true,   
   pdfusetitle=true,  
}

\title{Sequence-to-Sequence Spanish Pre-trained Language Models}

\name{Vladimir Araujo\textsuperscript{1},
Maria Mihaela Trusca\textsuperscript{1},
Rodrigo Tufi\~no\textsuperscript{2},
Marie-Francine Moens\textsuperscript{1}
}

\address{\textsuperscript{1}LIIR Lab, KU Leuven, Leuven, Belgium \\
\textsuperscript{2}IDEIAGEOCA, Universidad Politécnica Salesiana, Quito, Ecuador \\
\href{mailto:vgaraujo@uc.cl}{\texttt{vladimir.araujo@kuleuven.be}}\\}

\abstract{
In recent years, significant advancements in pre-trained language models have driven the creation of numerous non-English language variants, with a particular emphasis on encoder-only and decoder-only architectures. While Spanish language models based on BERT and GPT have demonstrated proficiency in natural language understanding and generation, there remains a noticeable scarcity of encoder-decoder models explicitly designed for sequence-to-sequence tasks, which aim to map input sequences to generate output sequences conditionally. This paper breaks new ground by introducing the implementation and evaluation of renowned encoder-decoder architectures exclusively pre-trained on Spanish corpora. Specifically, we present Spanish versions of BART, T5, and BERT2BERT-style models and subject them to a comprehensive assessment across various sequence-to-sequence tasks, including summarization, question answering, split-and-rephrase, dialogue, and translation. Our findings underscore the competitive performance of all models, with the BART- and T5-based models emerging as top performers across all tasks. We have made all models publicly available to the research community to foster future explorations and advancements in Spanish NLP: \url{https://github.com/vgaraujov/Seq2Seq-Spanish-PLMs}.
\\ \newline  }

\begin{document}

\maketitleabstract

\section{Introduction}
Spanish ranks among the most extensively used languages globally. This fact has captured the interest of the NLP community, prompting efforts towards resource development for this NLP domain. Consequently, a number of pre-trained language models tailored for Spanish have emerged in recent years, predominantly employing encoder-only \citep{canete2020spanish,SalasGrandury2022,make5010005} and decoder-only \citep{gut2021spanish} architectures. These models have demonstrated exemplary performance in natural language understanding across several downstream tasks \citep{canete2020spanish,araujo-etal-2022-evaluation}. Nevertheless, there has been limited advancement in addressing tasks revolving around generating new sentences depending on a given input, such as summarization, generative question answering, dialogue, or translation.

Encoder-decoder models primarily serve for addressing sequence-to-sequence tasks, and over recent years, numerous architectures have emerged.
The pre-training of these models is often based on the whole transformer architecture \citep{NIPS2017_3f5ee243} and entails more intricate learning objectives than those of encoder or decoder-only models individually. For instance, BART \citep{lewis-etal-2020-bart} is specifically trained to reconstruct text that has been intentionally corrupted, while T5 \citep{JMLR:v21:20-074} is designed to adeptly fill in missing sections of text, simulating a scenario where text spans have been omitted.
These models have been developed predominantly for the English language, and recent efforts have been made to pre-train them in languages other than English \citep{kamal-eddine-etal-2021-barthez}. Unfortunately, when it comes to the Spanish language, there is a notable scarcity of such models that may be valuable to the community.

In this paper, with the aim of democratizing sequence-to-sequence models for the Spanish NLP community, we introduce BARTO and T5S, which are the Spanish counterparts of the BART and T5 models.
These models are exclusively pre-trained on Spanish corpora, aligning with their self-supervised methodology.
We also introduce models in the style of BERT2BERT \citep{rothe-etal-2020-leveraging}, utilizing well-established BERT \citep{devlin-etal-2019-bert} and RoBERTa \citep{liu2020roberta} models for Spanish as baselines. 
Additionally, we curate a variety of Spanish sequence-to-sequence tasks, such as summarization, question answering, split-and-rephrase, dialogue, and machine translation, to assess our models comprehensively.

Our results demonstrate that all models perform well on the proposed benchmark tasks. Particularly, BARTO and T5S stand out in text generation tasks, especially with lengthy sequences, surpassing BERT2BERT-style models and their multilingual counterparts in many generative tasks. Furthermore, we evaluate the models' performance on discriminative tasks, such as sequence and token classification. While their performance is slightly behind encoder-only models in some tasks, they still offer very competitive results. To facilitate future research and practical applications in Spanish NLP, we have made these models available.

\section{Related Work}

\subsection{Language-specific Pre-trained Language Models}

Pre-trained language models represent a class of advanced language models trained through self-supervised learning on large text corpora, making them versatile for various applications. Notably, two prominent models are BERT \citep{devlin-etal-2019-bert}, an encoder-only model, and GPT \citep{radford2018,radford2019}, a decoder-only model. These models have established robust baselines for a wide range of NLP tasks in English.

Numerous language-specific BERT-based and GPT-based models have emerged in recent times. Examples include CamemBERT \citep{martin-etal-2020-camembert} tailored for French, RobBERT \citep{delobelle-etal-2020-robbert} designed for Dutch, FinBERT \citep{virtanen2019multilingual} for Finnish, GePpeTto \citep{demattei2020geppetto} for Italian, and several others. These models have consistently outperformed their multilingual counterparts, highlighting the value of their existence for language-specific tasks.

In the context of the Spanish language, we find BETO \citep{canete2020spanish} and ALBETO \citep{canete-etal-2022-albeto}, a BERT and ALBERT model, respectively, pre-trained on the SUC corpus \citep{canete2020spanish}. Regionalized BERT models for Spanish language variations \citep{Tellez2023} based on Twitter data.
BERTIN \citep{SalasGrandury2022}, a RoBERTa base model trained on the Spanish portion sampled from mC4 \citep{xue-etal-2021-mt5}. 
Furthermore, MarIA \citep{gut2021spanish} introduces a family of models, including RoBERTa and GPT-2 trained on the corpus crawled by the National Library of Spain. 
More recently, RigoBERTa \citep{Serrano2022RigoBERTaAS}, follows the DeBERTa \citep{He2020DeBERTaDB} architecture and was trained with several corpora, including OSCAR \citep{OrtizSuarezSagotRomary2019}, SUC, and mC4-es. 
Nevertheless, a notable gap exists in the availability of encoder-decoder models exclusively trained for Spanish.

\subsection{Sequence-to-Sequence Pre-trained Language Models}

A sequence-to-sequence model aims to map a fixed-length input with a fixed-length output where the length of the input and output may differ \citep{NIPS2014_a14ac55a}. 
It comprises an encoder, which concurrently processes the entire input sequence, and a decoder, which receives the representations computed by the encoder and generates the output sequence in an autoregressive manner.
These models have proven to be valuable in addressing tasks including translation, dialogue, question answering, and summarization.

Following the paradigm of pre-training with self-supervision, several models have been proposed. One of the first models is MASS \citep{pmlr-v97-song19d}, which uses a transformer to reconstruct an input sequence where a contiguous span of tokens is masked and mapped to a sequence consisting of the missing tokens. Later, T5 \citep{JMLR:v21:20-074} proposed a pre-train on a multitask combination of supervised and self-supervised tasks, the latter being a task to complete fill-in dropped-out spans of text from documents. BART \citep{lewis-etal-2020-bart} is slightly similar to T5 but only uses a self-supervised objective in which spans are masked from the input, but the complete output is predicted to improve the decoder’s language modeling ability.
Moreover, \citet{rothe-etal-2020-leveraging} proposed the utilization of encoder or decoder-only pre-trained checkpoints for initializing new encoder-decoder models, showcasing competitive performance compared to purely encoder-decoder pre-trained models.

More recently, there has been a notable surge in endeavors to deploy sequence-to-sequence models for languages beyond English. BART, for instance, has been released for many other languages, including French \citep{kamal-eddine-etal-2021-barthez}, Greek \citep{evdaimon2023greekbart}, Indic \citep{dabre-etal-2022-indicbart}, Arabic \citep{kamal-eddine-etal-2022-arabart} and various other languages \citep{Tran2022,fi15010015}. Furthermore, T5 has been pre-trained in Portuguese \cite{carmo2020ptt5}, Italian \citep{sarti2022it5}, Arabic \citep{nagoudi-etal-2022-arat5}, Indic \citep{aralikatte-etal-2023-varta}, among others.
While the aforementioned recent models encompass a broad range of languages, the availability of Spanish models remains limited.

\section{Sequence-to-Sequence Spanish Pre-trained Language Models}
In this section, we begin by presenting our data collection and preparation procedures for pre-training our models. Subsequently, we provide detailed descriptions of each model and outline the corresponding pre-training processes.

\subsection{Pre-training Data}

We employ the OSCAR 21.09 corpus \citep{abadji-etal-2022-towards}, which includes a deduplicated Spanish set of approximately 160GB of text. Furthermore, we utilize the mC4-es corpus \citep{xue-etal-2021-mt5}, specifically adopting the Gaussian perplexity sampling subset proposed by \citet{SalasGrandury2022}, which boasts an extensive 500GB text dataset and has demonstrated superior model consistency. Additionally, we incorporate SUC, the corpus utilized for pre-training BETO, comprising around 14GB of raw text from diverse sources. Note that we exclude Wikipedia data from SUC, instead opting for an updated Wikipedia dump\footnote{\href{https://dumps.wikimedia.org/eswiki/latest/}{https://dumps.wikimedia.org/eswiki/latest/}} of approximately 10GB.

As established by prior research \citep{liu2020roberta,JMLR:v21:20-074}, the corpus quality significantly impacts the outcomes of pre-training models. Consequently, we closely follow the preprocessing methodologies previously established for both English \citep{JMLR:v21:20-074} and Spanish models \citep{gut2021spanish,Serrano2022RigoBERTaAS}. Below, we describe the procedure:

\begin{enumerate}
\itemindent=-5pt
\itemsep0em 
\item \textbf{Document-level Formatting:} 
We ensure that all data adheres to a document-level format, which means that each instance is a document containing several contiguous coherent sentences. 
As demonstrated by \citet{liu2020roberta}, restricting sequences to come from a single document performs slightly better than packing sequences from multiple documents.
Furthermore, this causes the models to capture broad contextual dependencies.

\item \textbf{Data Filtering:} To enhance data quality, we employ straightforward and cost-effective filtering methods. We eliminate very short documents based on sentence and document length. We filter out text containing repeated characters or special characters not commonly used in Spanish. 
We exclude pages containing code and sensitive content.
Moreover, we utilize the fastText language identification model \citep{joulin-etal-2017-bag} to exclude documents classified with less than 98\% accuracy for Spanish text.
We set this threshold to ensure our corpus retains a small yet representative portion of other languages that are frequently interspersed in contemporary Spanish texts.

\item \textbf{Deduplication:} We employ a deduplication process across all corpora using the \texttt{text-dedup}\footnote{\href{https://github.com/ChenghaoMou/text-dedup}{https://github.com/ChenghaoMou/text-dedup}} library.
This generates a dataset smaller and faster to train on while potentially enhancing the resulting performance due to avoiding duplicated data \citep{lee-etal-2022-deduplicating}.
Due to its computational intensity, this step is performed after formatting and filtering. 

\item \textbf{Encoding Correction:} Some documents may have inconsistent encodings or exhibit encoding issues. To address this, we utilize the \texttt{ftfy}\footnote{\href{https://ftfy.readthedocs.io/}{https://ftfy.readthedocs.io/}} tool to rectify encoding mix-ups, ensuring UTF-8 encoding and NFKC normalization.
We carry out this process at the end of the pipeline because, while \texttt{ftfy} is highly effective in text correction, it can be computationally expensive for large corpora. Additionally, this guarantees that we have proper encoding after all the preceding steps.
\end{enumerate}

The resulting corpus size after preprocessing exceeds 120GB of uncompressed text, a scale similar to the one used for training RoBERTa and BART models \cite{liu2020roberta,lewis-etal-2020-bart}.

\subsection{BARTO Model}
BARTO follows the BART base architecture, which consists of an encoder and a decoder with 6 layers each. Also, it has 12 attention heads and 768 hidden dimensions in both the encoder and decoder. BARTO is pre-trained by denoising the corrupted input documents. As suggested by \citet{lewis-etal-2020-bart}, we use a combination of text infilling and sentence permutation transformations for robust performance, masking 30\% of tokens in each document and permuting all sentences.

We use \texttt{sentencepiece} \citep{kudo-richardson-2018-sentencepiece} to build a BPE tokenizer of 50,264 tokens. Furthermore, we rely on the \texttt{fairseq} library \citep{ott-etal-2019-fairseq} to perform the training. BARTO is pre-trained for 100,000 steps on 8 NVIDIA A100 GPUs with input texts of 1024 and a batch size of 2048. We use the Adam optimizer \citep{kingma:adam}, a warm-up of 10,000 steps, a dropout of 0.1, and FP16 to speed up training.

\subsection{T5S Model}

T5S follows the T5.1.1\footnote{\href{https://github.com/google-research/text-to-text-transfer-transformer/blob/main/released_checkpoints.md}{https://github.com/google-research/text-to-text-\\transfer-transformer/blob/main/released\_checkpoints.md}} base version of the T5 model, which includes some improvements. This model consists of an encoder and decoder with 12 layers, 12 attention heads, and 768 hidden dimensions each. Like T5.1.1, we pre-train only using the denoising objective by filling in dropped-out spans of text from documents.
As proposed by \citet{JMLR:v21:20-074}, we use a corruption rate of 15\% and an average span length of 3.

We use the \texttt{sentencepiece} library to build a unigram tokenizer of 32,000 tokens. 
We also rely on \texttt{nanoT5} \citep{nawrot2023nanot5}, which allows pre-training T5 models on a limited budget. Our T5S is pre-trained for 130,000 steps on 4 GPUs NVIDIA A100 with input texts of 1024 and a batch size of 320. Additionally, we use the AdamW optimizer \citep{loshchilov2018decoupled}, a warm-up of 10,000 steps, a dropout of 0, and BP16 to speed up training.

\subsection{BERT2BERT-style Models}
Our BERT2BERT-style models follow the procedure proposed by \citet{rothe-etal-2020-leveraging}, which consists of initializing encoder-decoder models with pre-trained encoder and/or decoder-only checkpoints. We use two configurations: BERT2BERT, which is an encoder initialized by a BERT-type checkpoint paired with a decoder initialized with the same checkpoint, and BERTShare, which is similar to BERT2BERT but the parameters between the encoder and decoder are shared.

We rely on the \texttt{transformers}\footnote{\href{https://huggingface.co/docs/transformers/model_doc/encoder-decoder}{https://huggingface.co/docs/transformers/\\model\_doc/encoder-decoder}} 
library \citep{wolf-etal-2020-transformers} to initialize models based on two well-known architectures. On the one hand, by using the BETO checkpoint
, we initialize a BETO2BETO and BETOShare model. On the other hand, by leveraging the RoBERTa-BNE checkpoint 
from MarIA, we initialize a RoBERTa2RoBERTa and RoBERTaShare. Note that these models do not need to continue pre-training but rather fine-tune them directly in downstream tasks. We will delve into this process in more detail in the following section.

\section{Evaluation}
In this section, we introduce the downstream tasks chosen to evaluate the performance of our sequence-to-sequence models. These datasets primarily consist of generative tasks, where both input and output sequence texts are provided. Furthermore, we describe some discriminative tasks we adopt as part of our assessment.
Finally, we provide details on the model fine-tuning process.

\subsection{Generative Tasks}

\paragraph{Abstractive Summarization}
Summarization involves creating a concise version of a document while retaining its key information.
We consider MLSUM \citep{scialom-etal-2020-mlsum} and WikiLingua \citep{ladhak-etal-2020-wikilingua} datasets to evaluate our models. 
On the one hand, MLSUM is a collection of newspaper articles with an average number of $\sim$900 tokens\footnote{Token counting is done using the BETO tokenizer.} and summaries of approximately $\sim$24 tokens.
On the other hand, WikiLingua consists of guide-based articles with around $\sim$500 tokens on average, and its summaries contain about $\sim$50 tokens.
Note that both datasets do not contain overly lengthy texts, particularly in the case of the summaries. This factor simplifies the generative process significantly.

\paragraph{Long-form Abstractive Summarization}

A more challenging task of abstractive summarization is the processing and/or generation of long-form text.
We use XL-Sum \citep{hasan-etal-2021-xl}, a dataset with lengthy articles, and EUR-Lex-Sum \citep{aumiller-etal-2022-eur}, a dataset with long summaries.
XL-Sum \citep{hasan-etal-2021-xl} contains long news articles with about $\sim$1200 tokens on average and short summaries of about $\sim$40 tokens. In contrast, EUR-Lex-Sum is a dataset of legal documents with an average length of $\sim$19000 tokens and lengthy summaries of approximately $\sim$1200 tokens. These datasets present more significant challenges compared to the previous ones. This is due to the more intensive nature of the encoding and generative process, which demands models capable of processing lengthy sequences and capturing extended dependencies to produce high-quality summaries.

\paragraph{Split and Rephrase}

The split-and-rephrase task assumes rewriting the content of a long sentence into shorter and less verbose sentences. We use the Spanish subset of the BiSECT dataset \citep{kim-etal-2021-bisect} to evaluate this task, which contains about 290,000 instances. 
The average number of tokens within the input sentences is approximately $\sim$51, while after rephrasing into two sentences, the average increases to $\sim$75 tokens across a pair of sentences. Following \citet{kim-etal-2021-bisect}, we use a special token to separate split sentences during model fine-tuning. Specifically, we use \texttt{<s>} (\texttt{[CLS]} in the case of BERT2BERT-style models) since it provides the best performance.

\paragraph{Generative Question Answering}
This task focuses on generating an abstractive answer to a given question from a provided passage.
To the best of our knowledge, there is currently no dataset tailored for abstractive question answering in Spanish. In line with prior work \citep{JMLR:v21:20-074}, we utilize use span-based question answering datasets to train the models to generate the correct answers rather than predicting the specific token positions of the answer.
We rely on MLQA~\citep{lewis2019mlqa} and SQAC~\citep{gut2021spanish} datasets for this evaluation. MLQA presents a collection of parallel multi-lingual articles extracted from Wikipedia and offers a development set and test set professionally translated into Spanish. Unlike MLQA, SQAC was proposed exclusively for Spanish evaluation and contains articles extracted from Spanish sources.
Following the BART fine-tuning procedure \citep{lewis-etal-2020-bart}, models generate answers conditioned on the concatenation of questions and supporting documents.

\paragraph{Dialogue}
The dialogue response generation task aims to produce an appropriate and coherent response based on the dialogue context.
We employ the Spanish partition of the MIAM dataset \citep{colombo-etal-2021-code}, a benchmark comprising dialogue act corpora. 
This dataset proposed initially to identify the specific act that a speaker performs is not the task we want to address in this work.
Therefore, we adapt the data to suit our evaluation needs. Following previous work \citep{zhou-etal-2021-generation,zhang-etal-2020-dialogpt}, we focus solely on the utterances exchanged between the speakers to create pairs of dialogue context and responses.

\paragraph{Machine Translation}
The objective of this task is to translate a sentence from a source language to a different target language.
We rely on Fapesp-v2 \citep{aziz-specia-2011-fully} and WMT13 \citep{bojar-etal-2013-findings}, well-known machine translation benchmarks that encompass Spanish-language data.
On the one hand, Fapesp-v2 is a Portuguese $\leftrightarrow$ Spanish parallel corpora crawled from a Brazilian magazine with about $\sim$150,000 examples of training sets and $\sim$1,300 examples of development and test sets.
On the other hand, WMT13 includes an English $\leftrightarrow$ Spanish subset comprising approximately 15 million training instances, which is extensive for conducting small-scale experiments. Therefore, we opted to work with a randomly sampled subset of 600,000 examples, a size similar to that used for fine-tuning English models \citep{lewis-etal-2020-bart}. For evaluation, we use the newstest2012 and newstest2013 sets, each containing $\sim$3000 examples.

\subsection{Discriminative Tasks}

\paragraph{GLUES}
We rely on the GLUES, a benchmark introduced initially by \citet{canete2020spanish}. We focus our evaluation on sequence classification tasks: MLDoc, PAWS-X, and XNLI.
MLDoc revolves around the classification of long documents into four distinct categories. PAWS-X, on the other hand, centers on the identification of sentence paraphrases. Lastly, XNLI consists of predicting whether a premise logically entails a given hypothesis.
Additionally, we use semantic similarity and relatedness tasks, including STS-es \cite{gut2021spanish} and SemRel2024 \cite{ousidhoum2024semrel2024}. STS-es entails evaluating the similarity between two text segments by assigning a score from 1 to 5. SemRel2024 assesses the degree of semantic textual relatedness between two sentences, assigning a score between 0 and 1.

\paragraph{SQAC}
In the case of token-level classification, we rely on the SQAC dataset \citep{gut2021spanish}. This corpus is a compilation of question-answer pairs extracted from Wikipedia articles. Within SQAC, the primary task involves predicting the specific span within the text that corresponds to the answer to a given question. This is achieved by indicating the start and end positions of this answer span within the text. As a common practice, the input to the model during fine-tuning is concatenating the question and the contextual text.

\begin{table*}[]
\centering
\small
\begin{tabular}{lcccccc}
\hline
 & \multicolumn{3}{c}{\textbf{MLSUM}} & \multicolumn{3}{c}{\textbf{WikiLingua}} \\
 & R1         & R2        & RL        & R1          & R2          & RL          \\
 \hline
 BETO2BETO   & 28.46/28.09      & 10.87/10.34     & 22.89/22.51     & 38.02/37.92       & 17.57/17.43       & 29.38/29.24       \\
 BETOShare & 28.51/27.84      & 10.90/10.19     & 22.99/22.30     & 37.74/37.68       & 17.41/17.29       &     29.19/29.03   \\
 RoBERTa2RoBERTa   & 27.94/27.69      & 9.66/9.25       & 21.92/22.07     & 35.68/35.58       & 14.53/14.51       & 26.49/26.37       \\
 RoBERTaShare & 28.43/27.86      & 10.17/9.53      & 22.54/21.92     & 35.83/35.70       & 14.95/14.75       &     26.85/26.62   \\
 \hline
 mBART-large   & 29.11/28.56      & 10.85/10.21     & 22.62/22.00   & 41.15/40.98       & 20.44/20.40       & 33.24/33.01       \\
 mT5     & 29.33/28.69      & 11.53/10.89     & 23.91/23.24     & 37.27/37.20       & 18.57/18.48       & 31.39/31.25       \\
 \hline
 BARTO   & 29.65/29.12      & 10.96/10.32     & 22.71/22.18     & 39.48/39.37       & 19.65/19.46       & 32.33/30.63       \\
 T5S    & 30.14/29.44      & 12.27/11.56     & 24.62/23.88     & 39.63/39.53       & 20.42/20.27       & 33.48/33.20       \\
\hline       
\end{tabular}
\caption{
Summarization task results on the development \textbf{/} test sets for all models using ROUGE metric.}
\label{tab:sum}
\end{table*}

\begin{table*}[]
\small
\centering
\begin{tabular}{lcccccc}
\hline
 & \multicolumn{3}{c}{\textbf{XLSum}} & \multicolumn{3}{c}{\textbf{EUR-Lex-Sum}} \\
 & R1         & R2        & RL        & R1          & R2          & RL          \\
 \hline
 BETO2BETO   & 28.76/28.88      &  8.92/9.02     & 20.85/21.03     & 42.76/43.46       & 13.89/14.17       &     22.77/23.07       \\
 BETOShare & 28.96/29.24      &  9.17/9.22     & 21.08/21.27     & 41.66/42.76       & 13.42/13.73       &     22.66/22.95   \\
 RoBERTa2RoBERTa   & 26.92/27.22      &  6.98/7.32     & 19.01/19.23     & 44.70/45.63       & 14.58/14.93       &     22.86/23.06       \\
 RoBERTaShare & 26.89/27.08      &  6.99/7.15     & 18.94/19.11     & 44.24/44.22       & 13.84/13.90       &     22.55/22.65   \\
 \hline
 mBART-large   & 31.56/31.76      &  10.94/10.89   & 22.26/22.27     & 68.20/67.37       & 52.89/51.45       &     58.05/56.35   \\
 mT5   & 28.54/28.59      &  10.18/10.28   & 21.42/21.49     & 65.66/64.20       & 50.32/48.88       &     56.07/53.95   \\
 \hline
 BARTO   & 31.02/31.26      &  10.68/10.72   & 21.96/23.81     & 66.49/65.91       & 49.99/48.39       &     56.01/54.15   \\
 T5S    & 30.59/30.80      &  11.99/12.14   & 23.38/23.48     & 64.94/64.61       & 49.84/49.14       &     55.45/54.56   \\
\hline       
\end{tabular}
\caption{
Long-form summarization task results on the development \textbf{/} test sets using ROUGE metric.}
\label{tab:xsum}
\vspace{-0.25cm}
\end{table*}

\begin{table*}[]
\centering
\small
\begin{tabular}{lcccccc}
\hline
 & \multicolumn{3}{c}{\textbf{SQAC}} & \multicolumn{3}{c}{\textbf{MLQA}} \\
 & R1         & R2        & RL        & R1          & R2          & RL          \\
 \hline
 BETO2BETO   & 23.84/24.19      & 10.89/11.10     & 22.87/23.14     & 35.01/33.10       & 23.91/22.29       & 34.36/32.37       \\
 BETOShare & 28.61/29.22      & 15.08/15.52     & 27.69/28.17     & 33.77/33.37       & 23.61/23.24       & 33.08/32.63       \\
 RoBERTa2RoBERTa   & 25.42/24.86      & 12.43/12.16     & 24.53/23.78     & 32.21/31.01       & 20.50/18.90       & 31.58/30.12       \\
 RoBERTaShare & 33.21/33.28      & 20.37/20.52     & 32.52/32.45     & 29.21/27.78       & 16.65/16.48       &     28.39/26.69   \\
 \hline
 mBART-large  & 70.97/71.40      & 52.25/53.08     & 70.80/71.24    & 62.55/59.66       & 37.87/34.73       & 62.39/59.50       \\
 mT5    & 75.14/74.11      & 56.03/55.77     & 75.07/73.98    & 70.93/69.49       & 46.32/43.42       & 70.80/69.36       \\
 \hline
 BARTO   & 77.92/77.00      & 58.88/58.45     & 77.78/76.87     & 68.69/66.45       & 44.08/40.87       & 68.57/66.34       \\
 T5S    & 80.68/78.80      & 60.39/59.33     & 80.64/78.64     & 72.02/70.45       & 47.53/44.37       & 71.84/70.32       \\
\hline      
\end{tabular}
\caption{
Generative question answering task results on the development \textbf{/} test sets using ROUGE.}
\label{tab:gqa}
\vspace{-0.25cm}
\end{table*}
\subsection{Fine-tuning}

We follow the fine-tuning procedures proposed for the English version models \citep{lewis-etal-2020-bart,JMLR:v21:20-074}. 
Because BARTO and T5S have an autoregressive decoder, they can be directly fine-tuned for sequence generation tasks. Specifically, their encoders take a complete input, and then their decoders generate a target output autoregressively.

For BETO2BETO and similar models, we initialize a transformer with the BETO checkpoint in both the encoder and decoder. 
Note that the decoder has cross-attention layers that are randomly initialized since BETO does not have these parameters.
Subsequently, we fine-tune these models following the same procedure as BARTO and T5S.

We fine-tune the models on an RTX 3090 GPU for each task using the \texttt{transformers} library implemented in PyTorch.
For a fair comparison, we use the same hyperparameters with the exception of the batch size, learning rate, and the number of training epochs. 
The optimal settings may depend on the task, therefore we consider a hyperparameter sweep with batch size $\in$ \{4, 8, 16\}, learning rate (AdamW) $\in$ \{3e -5, 5e-5\}, and epochs $\in$ \{3, 6\}.

\subsection{Multilingual Baselines}

To complement our evaluation, we include multilingual sequence-to-sequence baselines for further comparison. We employ mT5-base \cite{xue-etal-2021-mt5}, a model trained on 101 languages. While there is no base version of mBART directly comparable to our models, we include a mBART-large version for comparison purposes. Specifically, we use mBART-50 \cite{tang2020multilingual}, which includes both Spanish and Portuguese, essential for our experiments.
We fine-tune these models following the same procedure as for BARTO and T5S.

\section{Results}
\label{results}
This section presents the results achieved after fine-tuning all the models on the downstream tasks. We evaluate them using conventional metrics, adhering to established methodologies in previous work \cite{lewis-etal-2020-bart,JMLR:v21:20-074}.

\subsection{Generative Tasks}
\label{results:gen}

\paragraph{Abstractive Summarization}

Table~\ref{tab:sum} presents a comparison of the results achieved by all the models on the MLSUM and WikiLingua tasks, measured in terms of ROUGE metrics \citep{lin-2004-rouge}. 
Specifically, the T5S model shows the highest performance with an average of 26.54 ROUGE among the Spanish models across all tasks, while BARTO is second with a difference with an average of 25.49 ROUGE.
As for the BERT2BERT style models, they are all outperformed. BETO2BETO is the one that offers the best performance among its group of models, with an average of 24.39 ROUGE.
Notably, T5S and BARTO outperform their multilingual counterparts, with the exception of mBART on WikiLingua, which slightly outperforms T5S by 0.45 ROUGE.

\paragraph{Long-form Abstractive Summarization}

The results for long-form summarization are presented in Table~\ref{tab:xsum}. 
In XLSum, T5S slightly outperforms BARTO, averaging 22.06 and 21.58 ROUGE, respectively. Furthermore, BETOShare achieves the best performance of its group of models with an average of 19.82 ROUGE.
Regarding EUR-Lex-Sum, BARTO outperforms T5S, averaging 56.82 and 56.42 ROUGE, respectively.
Notably, the difference between BARTO and T5S in comparison to the BERT2BERT-style models is significant, being BETO2BETO the best performer with an average of 26.69 ROUGE. 
These results reflect the unique requirements of EUR-Lex-Sum, which involve processing lengthy articles and generating extensive summaries. BART and T5S show better performance in this task, mainly because they leverage document formatting during pre-training. This strategy enhances their ability to handle long sequences, especially when dealing with extended documents.
Finally, the multilingual models slightly outperform BARTO and T5S only in the case of EUR-Lex-Sum. This may be due to the diverse range of text types encountered by multilingual models during pre-training, including legal documents.
\hyperref[appendix:b]{Appendix A} presents additional experiments exploring the potential of BARTO as a Longformer \cite{beltagy2020longformer} for this task.

\paragraph{Generative Question Answering}

Table~\ref{tab:gqa} presents the results of the generative question answering tasks in terms of ROUGE scores.
Although SQAC and MLQA were originally designed as discriminative tasks, our results indicate that they serve as a suitable benchmark for generative question answering.
BARTO and T5S show the best performance in all tasks, with T5S being the best, reaching 67.92 ROUGE on average.
In this task, the performance difference between these two models is notably significant compared to BERT2BERT-style models, with BETOShare as the top performer of the group with an average of 27 ROUGE.
This difference may stem from the self-supervised objective of BART and T5, enabling better transfer learning to this task compared to others \citep{JMLR:v21:20-074}.
Significantly, in these tasks, both BARTO and T5S outperform their multilingual counterparts, including the large-sized mBART. These results highlight the superior reading comprehension capabilities of our models.

\paragraph{Split and Rephrase}

Table~\ref{tab:sandr} presents the comparison of models on the BiSECT dataset, with evaluations based on SARI \citep{xu-etal-2016-optimizing} and BLEU \citep{post-2018-call} scores. T5S achieves the highest scores for both metrics, averaging 56.37 SARI and 43.27 BLEU. While BARTO secures the second-highest BLEU score, its performance in SARI falls short. In particular, RoBERTaShare has the best performance among BERT2BERT style models, averaging 51.29 SARI but outperformed by BARTO in BLEU. T5S's success could be attributed to its ability to generate sequences that closely resemble the input's word order, a trait closely related to SARI's evaluation criteria. Additionally, T5's span-filling objective may facilitate sentence splitting, contributing to its overall performance boost. Lastly, T5S outshines its multimodal counterpart, while BARTO performs comparably to mBART-large.

\begin{table}[]
\small
\centering
\begin{tabular}{lcc}
\hline
 & \multicolumn{2}{c}{\textbf{BiSECT}} \\
 & SARI  & BLEU  \\
\hline
BETO2BETO   & 49.45/49.27 & 37.79/37.14 \\
BETOShare & 49.72/49.37 & 38.38/37.62 \\
RoBERTa2RoBERTa   & 50.98/50.56 & 36.00/35.22 \\
RoBERTaShare & 51.49/51.09 & 37.19/36.16 \\
\hline
mBART-large   & 50.20/50.13 & 39.78/39.24 \\
mT5    & 55.91/55.74 & 43.39/42.62 \\
\hline
BARTO   & 50.45/50.13 & 39.48/38.97 \\
T5S    & 56.55/56.19 & 43.73/42.81 \\
\hline       
\end{tabular}
\caption{
Split-and-rephrase task results on the development \textbf{/} test sets for all models using SARI and BLEU metrics.}
\label{tab:sandr}
\end{table}

\paragraph{Dialogue}

Table~\ref{tab:dialogue} compares the models for dialogue generation based on F1 and METEOR scores \citep{banerjee-lavie-2005-meteor}. BARTO leads with 34.30 F1, closely trailed by BETOShare at 32.45 F1. Conversely, for the METEOR score, BETOShare outperforms with 27.33 METEOR, followed by BARTO at 26.95 METEOR.
Interestingly, T5S ranks fourth with averages of 28.80 F1 and 22.24 METEOR. T5S excels in tasks with high overlap between input and output, such as summarization or split-and-rephrase, which may explain its lower performance in dialogue generation where such overlap is reduced. Notably, multilingual models exhibit similar behavior to our models, with mBART outperforming mT5. This outcome suggests multilingual models have the broader knowledge required for dialogue tasks, giving them a slight edge over BARTO and T5S.

\begin{table}[]
\small
\centering
\begin{tabular}{lccc}
\hline
 & \multicolumn{2}{c}{\textbf{MIAM}} \\
 & F1 & METEOR \\
\hline
BETO2BETO   & 32.23/31.84 & 27.24/25.79 \\
BETOShare & 32.83/32.08 & 28.57/26.10 \\
RoBERTa2RoBERTa   & 19.63/14.05 & 18.73/12.49 \\
RoBERTaShare & 22.18/16.40 & 21.11/14.96 \\
\hline
mBART-large & 37.11/36.65 & 27.12/26.03 \\
mT5 &  29.41/30.90 &  19.00/19.68 \\
\hline
BARTO   & 34.19/34.42 & 27.66/26.24 \\
T5S & 28.37/29.27 & 22.72/21.76 \\
\hline       
\end{tabular}
\caption{
Dialogue response generation task results on the development \textbf{/} test sets for all models using F1 and METEOR metrics.}
\label{tab:dialogue}
\vspace{-0.25cm}
\end{table}

\begin{table*}[]
\small
\centering
\begin{tabular}{lccccc}
\hline
 & \multicolumn{2}{c}{\textbf{Fapesp-v2}} & \multicolumn{2}{c}{\textbf{WMT13}} \\
 & \textbf{PT} $\rightarrow$ \textbf{ES} & \textbf{ES} $\rightarrow$ \textbf{PT} & \textbf{EN} $\rightarrow$ \textbf{ES} & \textbf{ES} $\rightarrow$ \textbf{EN} \\
\hline
mBERT2mBERT   & 55.14/61.21 & 54.90/60.01 & 25.58/22.29 & 22.30/19.62 \\
mBERTShare  & 54.21/60.75 & 54.71/59.86 & 25.18/21.58 & 22.47/19.68 \\
\hline
mBART-large     & 65.64/70.89 & 67.16/71.44 & 29.33/29.37 & 32.28/30.71 \\
mT5     & 66.58/72.29 & 68.09/71.02 & 31.05/28.89 & 29.68/28.46 \\
\hline
BARTO     & 65.84/71.06 & 66.98/70.25 & 30.22/27.41 & 30.22/28.64 \\
T5S     & 66.37/71.85 & 68.65/72.40 & 31.99/29.37 & 31.10/29.64 \\
\hline       
\end{tabular}
\caption{
Machine translation task results on the development \textbf{/} test sets of Fapesp-v2 and newstest2012~ \textbf{/} newstest2013 sets of WMT13 using BLEU metric.}
\label{tab:mt}
\end{table*}

\begin{table*}[]
\centering
\small
\begin{tabular}{lcccccc}
\hline
 & \textbf{MLDoc} & \textbf{PAWS-X}  & \textbf{XNLI} & \textbf{STS-es} & \textbf{SemRel2024} & \textbf{SQAC} \\
 & Accuracy & F1 & Accuracy & Combined & Combined & F1 \\
\hline
BETO    & 96.20/95.77 & 86.39/87.98 & 81.12/80.89 & 89.50/81.02 & 74.20/75.47 & 79.85/79.15 \\
RoBERTa-BNE    & 95.80/95.92 & 88.34/89.76 & 80.68/80.63 & 89.37/80.46 & 71.39/75.09 & 80.16/80.39 \\
\hline
BARTO     & 96.70/96.20 & 87.39/88.92 & 81.81/79.78 & 86.39/81.73 & 72.14/75.61 & 79.22/79.09 \\
T5S$^\dagger$ & 96.90/96.63 & 88.61/88.58 & 80.92/80.56 & 86.64/75.65 & 63.84/66.60 & 58.46/62.29 \\
\hline       
\end{tabular}
\caption{
GLUES and SQAC tasks results on the development \textbf{/} test sets for all models. The combined metric represents the averaged Pearson and Spearman metrics. $^\dagger$ indicates that T5S uses ``text-to-text'' format \citep{JMLR:v21:20-074} to solve the tasks.}
\label{tab:glues}
\vspace{-0.25cm}
\end{table*}

\paragraph{Machine Translation}

BARTO and T5S are pre-trained with documents that comprise at least 98\% accuracy for Spanish language prediction. We hypothesize that this threshold allows our models to acquire knowledge of other languages often found alongside Spanish, allowing them to perform the translation task.
Therefore, we decide to fit our models directly in their original form, according to the procedure of previous tasks.
This contrasts with the approach suggested by \citet{lewis-etal-2020-bart}, which introduces new parameters to facilitate the adaptation of BART to a new language.

For better comparison, we employ multilingual BERT checkpoints to initialize mBERT2mBERT. We evaluate all the modes using the BLEU score. In our experiments, we found that maintaining a substantial batch size (e.g., 384) and limiting the number of fine-tuning steps positively impacts the resulting performance of both BARTO and T5S, but not mBERT2mBERT.

Table~\ref{tab:mt} presents the experimental results. Both BARTO and T5S show similar performance in translating from Portuguese to Spanish (Fapesp-v2 PT$\rightarrow$ES) and English to Spanish (WMT13 EN$\rightarrow$ES). Additionally, the top-performing BERT2BERT-style model is mBERT2mBERT.
Note that the BLEU score is higher for Fapesp-v2, and the performance gap of BARTO and T5S versus mBERT2mBERT is also more significant compared to WMT13. Our hypothesis is that this difference arises not only from the models exposed to the Portuguese text but also from the shared linguistic roots with Spanish. In fact, when analyzing the tokenizer, we observe Portuguese diacritics (e.g., ç, ã, ü).

Table~\ref{tab:mt} also shows the results of the experiments with Spanish as the source language for translation (Fapesp-v2 ES$\rightarrow$PT and WMT13 ES$\rightarrow$EN). Intuitively, one might expect our models to perform better when tasked with generating Spanish. However, they demonstrate competitive performance even when generating Portuguese and English.
Surpassing mBERT2mBERT, the best BERT2BERT-style model, BARTO and T5S exhibit more competitive performance for Fapesp-v2 and WMT13. Interestingly, the BLEU score is slightly higher than when generating Spanish text. We hypothesize that this is due to the effectiveness of the encoding representation of the Spanish input for conditionally generating quality text.

Finally, we find that BARTO and T5S perform similarly to their multilingual versions. These findings underscore that a sequence-to-sequence model, pre-trained from scratch in Spanish with a minimal amount of input from another language, can effectively address translation tasks.

\subsection{Discriminative Tasks}

The results for the discriminative tasks are presented in Table~\ref{tab:glues}. 
For all sequence classification tasks, we adopt the input formatting conventions established in previous work \citep{canete2020spanish,gut2021spanish}. Subsequently, we utilize either the \texttt{<s>} or \texttt{[CLS]} token to perform the classification of the input text when employing BERTO, RoBERTa, or BARTO.
However, since this special token is missing for T5S, we fine-tune the model to generate the class label following the ``text-to-text'' format proposed by \citet{JMLR:v21:20-074}.

In the case of MLDoc, our findings indicate that BART and T5S exhibit superior performance, which can likely be attributed to their proficiency in handling lengthy sequences.
Regarding PAWS-X and XNLI, our results show that BART and T5S, while not securing always the top positions, still exhibit competitive performance. Finally, in the case of STS-es and SemRel2024, we find that encoder-only models perform the best possible because of their specialization on sentence representation. Our results align with the behavior of English models, where BART and T5 tend to slightly lag behind the state-of-the-art in sentence-paired classification tasks \citep{lewis-etal-2020-bart,JMLR:v21:20-074}.

In the token classification task, we follow the standard span-based question answering approach, except for T5S, which remains operating as a generative model, as previously outlined. Consistent with our sequence classification findings, we observe a slight performance advantage for encoder-only models. Interestingly, T5S shows the lowest performance in this scenario. This discrepancy could be attributed to the F1 metric's suitability for discriminative assessment, potentially overlooking T5S's generative capabilities. However, it is worth noting that T5S has demonstrated its ability to generate accurate answers in Section~\ref{results:gen}.

\begin{table*}[]
\centering
\small
\renewcommand{\arraystretch}{1.25} 
\begin{tabular}{l m{13cm}}
\hline
\multicolumn{2}{l}{\textbf{Long-form Summarization: WikiLingua}} \\
\hline
Source    & En caso de que hayas planificado con anticipación, esto te será fácil. Luego, regresa al ensayo cuando hayan pasado uno o dos días y revísalo [...]. Lee tu ensayo en voz alta, leyendo exactamente lo que haya en la hoja [...].  Podría serte de ayuda que imprimas el borrador [...]. Al trabajar con una copia física, te obligas a prestar atención de una forma nueva [...]. \\
\hline
Target    & Tómate uno o dos días sin mirar al ensayo. Lee el borrador. Revisa en donde sea necesario. Corrige con cuidado. \\
\hline
T5S       & Regresa al ensayo al día siguiente. Revisa el ensayo. Edita el ensayo. Revisa el ensayo \\
BARTO     & Revisa tu ensayo antes de enviarlo. Lee tu ensayo de principio a fin. Revisa tu ensayo en voz alta. Imprime y edita tu ensayo. Listo.   \\
BETO2BETO & Regresa el ensayo a un momento conveniente. Revisa tu ensayo de principio a fin. Corrige tu ensayo en voz alta. Haz una copia física de tu ensayo.  \\
\hline
\multicolumn{2}{l}{\textbf{Generative Question Answer: SQAC}} \\
\hline
Source & Anillo halo El anillo halo es el más interno y grueso de los anillos de Júpiter. Su borde exterior coincide con el interior del anillo principal aproximadamente a un radio de 122 500 km [...].\\
Question    & ¿Cuál es el anillo más interno y grueso de los anillos de Júpiter? \\
\hline
Target & halo \\
\hline
T5S       & El anillo halo \\
BARTO     &  anillo halo   \\
RoBERTaShare & el anillo halo: Iris (Aright (Award (Aver \\
\hline
\multicolumn{2}{l}{\textbf{Dialogue Generation: MIAM}} \\
\hline
Source  & 
 [...] . Hay varios trenes . El primero sale a la una y 50 de la madrugada y el último a las 7 de la tarde . Desea algo más ? . Sí , me gustaría obtener el tipo de tren para ir a Barcelona el próximo sábado día uno . Lo siento . No he entendido bien . puede repetir ? .\\
\hline
Target & Sí , me gustaría obtener el tipo de tren para ir a Barcelona el próximo sábado día uno . \\
\hline
T5S       & Me gustaría obtener el tipo de tren para ir a Barcelona el próximo sábado día uno. \\
BARTO     &  Me gustaría obtener el tipo de tren para ir a Barcelona el próximo sábado día uno.  \\
BETOShare & Sí, me gustaría saber el horario de trenes de Barcelona a Zaragoza el próximo sábado. \\
\hline
\multicolumn{2}{l}{\textbf{Machine Translation: WMT13 EN$\rightarrow$ES}} \\
\hline
Source  & Information entry has improved (the system receives more data), but output still needs to get better.\\
\hline
Target & La introducción de información ha mejorado (el sistema recibe más datos), y la salida todavía no es muy buena. \\
\hline
T5S       & La entrada de información ha mejorado (el sistema recibe más datos), pero la salida todavía necesita mejorarse. \\
BARTO     &  La entrada de información ha mejorado (el sistema recibe más datos), pero todavía es necesario mejorar la salida.  \\
mBART2mBART & La entrada de información ha mejorado ( el sistema recibe más datos ), pero la salida todavía necesita mejorarse mejorando la información. \\
\hline

\end{tabular}
\caption{Predictions generated by BARTO, T5S, and BERT2BERT style models for qualitative comparison in abstractive summarization, generative question answering, dialogue, and machine translation tasks.}
\label{tab:qualitative}
\end{table*}

\section{Qualitative Analysis}
Quantitative evaluation of BARTO and T5S (Section~\ref{results}) demonstrates an improvement in performance compared to BERT2BERT style models. To gain deeper insights into the behavior of these two models, we conducted a qualitative analysis of their predictions focused on abstractive summarization, generative question answering, translation, and dialogue tasks. Table~\ref{tab:qualitative} shows the selected examples generated by BARTO and T5S, highlighting their respective strengths and weaknesses compared to a BERT2BERT-style baseline.

In the context of the summarization task, both BARTO and T5S exhibit the ability to generate coherent summaries that align with the source content. BARTO's output tends to closely match the target text, maintaining a high textual similarity. 
Additionally, T5S works similarly, but in this specific example, we see a hallucination in its summary, "Escribe tu ensayo en voz alta", which is not present in the source or target text. This could explain the superiority of BART in this task compared to T5S.

Shifting to the generative question answering task, it becomes evident that T5S and BARTO excel in generating more comprehensive answers. While RoBERTaShare initially generates similar responses, it occasionally introduces superfluous words that are unrelated to the original question.

Regarding the automatic translation task, both BARTO and T5S generate good translations that are not identical to the target but are completely coherent and valid. However, mBERT2mBERT hallucinates generating "mejorando la información", something that is not in the source text.

Finally, in the case of dialogue generation, both BARTO and T5S create responses that are almost perfectly aligned with the target response. On the other hand, BETOShare hallucinates and mixes information specified in the source text or dialogue.

These illustrative examples highlight the skills of BARTO and T5S, showcasing their strong grasp of the Spanish language and their ability to generate natural and contextually relevant responses.

\section{Conclusions}
This work lays the foundation for future research in encoder-decoder architectures within the Spanish language domain. We present BART, T5, and BERT2BERT-style models, all exclusively pre-trained in Spanish. These models are accompanied by a diverse range of generative tasks intended to facilitate comprehensive evaluation. Our evaluation has demonstrated the effectiveness of these models in addressing these challenges, with BARTO and T5S emerging as the top performers.

As we look forward, there is potential to fill the gaps in sequence-to-sequence tasks by building specialized datasets. Additionally, a thorough comparative analysis involving monolingual and multilingual models, akin to the study by \citet{PLN6487}, could offer valuable insights into their strengths and limitations. Lastly, we envision the pre-training of larger-scale language models, such as Llama \citep{touvron2023llama}, to pave the way for advancements in emerging areas like chatbots.

\section{Ethics Statement and Limitations}

Our work presents new language models pre-trained exclusively in Spanish.
The data used to train the models do not imply any violation of privacy.
The potential negative social impacts from this work are similar to any other NLP models. Language models could potentially be used to create malicious and biased systems.

In this paper, we introduce base-sized language models, which may not be suitable for tasks requiring advanced capabilities. The current preference for larger models, which exhibit improved performance and emerging capabilities, directs our future efforts toward the release of larger architectures.

For the generative question answering evaluation, we utilized span-based datasets. However, using these datasets generatively may lead the model to focus on reproducing exact information from the source text. Future efforts should prioritize creating and evaluating actual abstractive question answering datasets that present a more diverse range of answers beyond the input replication.

\section{Acknowledgements}

This work was funded by the European Research Council Advanced Grant 788506 and supported by the Google Cloud Research Credits program (GCP) and TPU Research Cloud program (TRC).

\nocite{*}
\section{Bibliographical References}\label{sec:reference}

\bibliographystyle{lrec-coling2024-natbib}
\bibliography{lrec-coling2024-example}


\clearpage
\onecolumn

\section*{Appendix A: Longformer-Encoder-Decoder for Spanish (LEDO)}
\label{appendix:b}

Transformer-based pre-trained language models face challenges in processing long sequences due to the quadratic scaling of their self-attention mechanism with sequence length. To address this issue, \citet{beltagy2020longformer} introduced Longformer, which combines windowed local-context self-attention with task-motivated global attention to alleviate the maximum input length limitation. An extension of Longformer, LED, adapts BART for this purpose. In our work, we explore the suitability of our BARTO model within this framework. We present LEDO, a model capable of processing sequences up to 16K tokens. Following the approach of \citet{beltagy2020longformer},  we build LEDO by leveraging the weights of BARTO and initializing its new position embedding matrix by repeatedly copying BARTO’s 1K position embeddings 16 times.

\begin{table*}[!htbp]
\small
\centering
\begin{tabular}{lcccccc}
\hline
 & \multicolumn{3}{c}{\textbf{XLSum}} & \multicolumn{3}{c}{\textbf{EUR-Lex-Sum}} \\
 & R1         & R2        & RL        & R1          & R2          & RL          \\
 \hline
 BARTO   & 31.02/31.26      &  10.68/10.72   & 21.96/23.81     & 66.49/65.91       & 49.99/48.39       &     56.01/54.15   \\
 LEDO    & 32.23/32.19      &  12.82/12.72   & 24.10/24.13     & 62.12/61.11       & 40.98/39.48       &     44.95/43.51   \\
\hline       
\end{tabular}
\caption{
Long-form summarization task results on the development \textbf{/} test sets using ROUGE metric.}
\label{tab:xsum-led}
\end{table*}

Table~\ref{tab:xsum-led} presents the results for long-form summarization using BARTO and LEDO. We maintain consistent setup and hyperparameters for running the LEDO experiments. Interestingly, LEDO demonstrates superior performance over BARTO in XLSum, with an average of 23.03\%, showcasing its capacity to handle lengthy inputs more effectively. However, in EUR-Lex-Sum, BARTO outperforms LEDO, which achieved an average performance of 48.69\%. This difference may be attributed to the use of the same hyperparameters for LEDO as those for BARTO, which might not be optimal. Further investigation is required to fully explore the potential of LEDO, which will be pursued in future work.

\end{document}